\def\year{2021}\relax
\documentclass[letterpaper]{article} 
\usepackage{aaai21}  
\usepackage{times}  
\usepackage{helvet} 
\usepackage{courier}  
\usepackage[hyphens]{url}  
\usepackage{graphicx} 
\usepackage{natbib}  
\usepackage{caption} 

\usepackage[switch]{lineno} 

\usepackage{csquotes}
\usepackage[normalem]{ulem} 
\usepackage[inline]{enumitem}
\usepackage{multirow}

\usepackage{cognac_v1.0.3}
\usepackage[inline]{enumitem}
\usepackage[lofdepth,lotdepth]{subfig}
\usepackage{pgf}
\usepackage{pgf-pie}
\usepackage{pgfplots}
\pgfplotsset{width=7cm,compat=1.8}
\usepackage{booktabs}
\usepackage[normalem]{ulem}

\usepackage{color}
\usepackage{soul}
\definecolor{pigment}{rgb}{0.2, 0.2, 0.6}

\definecolor{blue}{RGB}{0, 93, 170}			
\definecolor{darkgreen}{HTML}{3bb35b}

\title{Toward an Atlas of Cultural Commonsense for Machine Reasoning}

\author {
        Anurag Acharya,\textsuperscript{\rm 1}
        Kartik Talamadupula, \textsuperscript{\rm 2}
        Mark A Finlayson \textsuperscript{\rm 1}
        \\
}
\affiliations {
    \textsuperscript{\rm 1} 
    School of Computing and Information Sciences \\ 
    Florida International University \\
    \textsuperscript{\rm 2} IBM Research\\
    \texttt{\{aacharya, markaf\}@fiu.edu, krtalamad@us.ibm.com}
}

\begin{document}


\maketitle

\begin{abstract}

Existing commonsense reasoning datasets for AI and NLP tasks fail to address an important aspect of human life: cultural differences. We introduce an approach that extends prior work on crowdsourcing commonsense knowledge by incorporating differences in knowledge that are attributable to cultural or national groups. We demonstrate the technique by collecting commonsense knowledge that surrounds six fairly universal rituals---birth, coming-of-age, marriage, funerals, new year, and birthdays---across two national groups: the United States and India. Our study expands the different types of relationships identified by existing work in the field of commonsense reasoning for commonplace events, and uses these new types to gather information that distinguish the identity of the groups providing the knowledge. It also moves us a step closer towards building a machine that doesn't assume a rigid framework of universal (and likely Western-biased) commonsense knowledge, but rather has the ability to reason in a contextually and culturally sensitive way. Our hope is that cultural knowledge of this sort will lead to more human-like performance in NLP tasks such as question answering (QA) and text understanding and generation. 


\end{abstract}

\section{Introduction}

In the past few years, there have been major advancements in the field of question answering (QA) systems~\cite{gan2019improving, fan2019scoring, qu2019answer, zafar2020iqa}, in which researchers have looked at different ways in which these systems can be made more accurate and human-like in both their responses as well as their methodology. Incorporating commonsense knowledge and reasoning into NLP systems is one such area of recent focus \cite{tandon2018reasoning, tandon2018commonsense, trinh2018simple, merkhofer2018mitre}, and a large body of recent work has focused on the creation, curation, and use of large-scale commonsense knowledge bases and knowledge graphs~\cite{sap2019atomic,bosselut2019comet}. Importantly, these types of knowledge acquisition efforts have a long history in and have been of great use to a wide variety of AI systems~\cite{shi2017data,olteanu2017distilling,sap2019socialbias,liu2020k}.

The importance of commonsense knowledge bases and repositories is clear from the volume of recent work that makes use of resources such as ConceptNet~\cite{speer2020rcqa,speer2016conceptnet,speer2012conceptnet} to imbue NLP systems with worldly knowledge obtained from humans. A key recent contribution along these lines was ATOMIC~\cite{sap2019atomic}, which tackles the task of incorporating commonsense reasoning into NLP tasks by generating an atlas of ``if-then'' rules that taken together produce behavior akin to commonsense reasoning. Work such as ATOMIC and COMET~\cite{bosselut2019comet} has made commonsense knowledge more accessible to the current generation of NLP systems; the progress and pitfalls of this work have been cataloged recently~\cite{sap2020commonsense}. 

One glaring omission in all of this prior work has been the lack of focus on context-contingent aspects of commonsense knowledge; that is, most prior work views commonsense as a universal monolith. While some events included in prior work are not variable across groups----like reading a book or breaking a window, for instance---many events are variable, and here we focus on one highly relevant type of context-specific commonsense knowledge, namely cultural commonsense. Consisting of ritualistic, geographical, and social knowledge, cultural commonsense plays a large but hidden role in humans' day-to-day social interactions.

For example, let us consider a very simple social setting: {\em You are invited to a wedding. How long do you expect to be gone for, and how many people do you think will be there?} For most people in the United States or the wider Western world, the answer would probably be a few hours; probably half a day, starting in the early afternoon; and somewhere around a 100 people. However, for many people in India, the obvious answer is that you will probably have to lay aside several days for the whole event, and anywhere between several hundred to over a thousand people will attend. Such socially-conditioned knowledge is inherently obvious to people from the respective cultures, and hints at the differences in commonsense knowledge across cultural and social settings, particularly when it comes to ritualistic practices. 


We build upon prior work on systematizing commonsense knowledge for use in NLP tasks by demonstrating a proof-of-concept scheme for gathering {\em cultural commonsense} in a format similar to previous approaches like ATOMIC. Specifically, we start by surveying the extensive prior literature on cultural knowledge and ritual practices, and select a short list of six rituals to focus on for our study. We select two different national groups that are diverse in their ritualistic practices, and conduct a pilot experiment via a survey. We report on the results of the survey, and showcase what a truly cultural commonsense knowledge repository might look like. We hope that this work spurs future research on incorporating cultural and social commonsense knowledge into NLP systems across a wide range of tasks.


\section{Related Work}
\label{sec:related_work}
 
While the concept of incorporating cultural knowledge into commonsense is novel, there have been several previous attempts at laying the groundwork for it.
We build on the ATOMIC~\cite{sap2019atomic} system and knowledge repository, where crowdsourced commonsense information was used to build an atlas for if-then reasoning. ATOMIC builds a knowledge graph containing inferential knowledge regarding $24,000$ ``short events.'' The dataset is then used on the social question answering system SocialIQA~\cite{sap2019social}, which shows an increase in performance using the commonsense knowledge from ATOMIC.

Another prominent effort is the AI2 Reasoning Challenge (ARC)~\cite{clark2018think}. ARC consisted of a dataset of almost 8,000 science questions in English. This dataset was split into the {\em Easy} set and the {\em Challenge} set; the {\em Challenge} set consisted of questions that neither a retrieval-based algorithm nor a word co-occurrence algorithm were able to answer correctly. Later, \citet{boratko2018systematic} more precisely analyzed the ARC knowledge, defining seven knowledge types and nine reasoning types, as well as triple annotating 192 ARC questions. 


In addition to ATOMIC and ARC, there have been several other efforts in the field of commonsense reasoning. \citet{rashkin2018event2mind} built a system called Event2mind---a commonsense inference system on events, intents, and reactions. Furthermore, \citet{speer2016conceptnet} have built an updated version of ConceptNet~\cite{liu2004conceptnet}, a practical commonsense reasoning toolkit, which is now a multilingual graph of general knowledge. Another relevant work in the field is Webchild 2.0, a fine-grained commonsense knowledge distillation~\cite{tandon2017webchild}. Several of the other important pieces of work done in the field of commonsense have been reviewed by \citet{davis2015commonsense}, which lays out the uses, successes, challenges, approaches and possible future work in the field of commonsense reasoning. Apart from these applications, \citet{gordon2017formal} lays out a formal theory of commonsense psychology and how people assume others think, while \citet{lake2017building} put forward their argument as to how we can go about building machines that learn and think like people.

There have also been several efforts to facilitate Question Answering (QA) systems and commonsense knowledge by building datasets~\cite{talmore2018commonsense,dunn2017searchqa,joshi2017triviaqa}. An important one is the Stanford Question Answering Dataset (SQuAD)~\cite{rajpurkar2016squad}, consisting of 100,000+ questions and a reading comprehension dataset. They contrast three types of tasks: reading comprehension (RC; read a passage, select a span that answers); Open-domain QA (answer a question from a large set of documents); and Cloze datasets (predict a missing word in a passage). Similarly  \citet{sap2019social} introduces a benchmark dataset for social and emotional commonsense reasoning and the MCTest dataset~\cite{richardson2013mctest}, which consists of multiple-choice reading comprehension questions comprised of short (150--300 words) fictional stories which were targeted at seven-year olds.


\section{Cultural Knowledge and Rituals}
\label{sec:cultural_knowledge_rituals}


The primary focus of this work is to encode cultural knowledge as an essential part of commonsense knowledge. To do this, we need to first find ways to collect cultural knowledge across a variety of groups. This is not straightforward, since cultural knowledge varies dramatically. Prior literature from the study of cultural groups suggests that it is hard enough just to define ``culture'' given the many complexities and nuances of group differences; it is even harder to identify the knowledge that comes along with it and to differentiate across cultural groups. Furthermore, as cultural norms and practices are so varied across various groups, it is hard to find knowledge structures that can be clearly compared across different groups. The first task, therefore, is to find topics that are relatively common across different groups that can be compared in a like-for-like fashion.

\subsection{Rituals}
\label{subsec:rituals}

We first detail our search for cultural markers that can be consistently compared across different groups. After a survey of the extant literature, we decided to focus on {\em rituals} that are commonly found across cultures. There are several definitions of {\em ritual}, and the concept often comes entangled with religion and rites~\cite[p. 259--262]{braun2000guide}. For the purposes of this work, we define ritual as a ``culturally defined set of behavior''~\cite{leach1968runaway}.





While a detailed review of the theory of rituals is beyond the scope of this paper, we can list several reasons for focusing on rituals as indicators of cultures. First, rituals are well-studied; Scholars have identified not just specific rituals, but also the genres and types of rituals and their variations across cultures, etc.~\cite{durkheim2008elementary,turner1973symbols,bell1992ritual,bell1997ritual}. There is also a rich body of literature on the analysis of cultural practices for various rituals across several cultures, clearly indicating a strong concomitance between rituals and cultures~\cite{gray1979keep, smith1986unity, underhill2000analysis}. 




While there great variety in the conceptualization of ritual, the most important taxonomies of ritual show  basic agreement on their core categories. Rituals can be classified in several ways: 
one of the more widely accepted categorizations of rituals by \citet{bell1992ritual,bell1997ritual} presents a compromise of six categories that are not necessarily mutually exclusive. These categories are: (i) {\it rites of passage}, a.k.a. {\it life-cycle rites}; (ii) {\it calendrical and commemorative rites}; (iii) {\it rites of exchange and communion}; (iv) {\it rites of affliction}; (v) {\it rites of feasting, fasting, and festivals}; and (vi) {\it political rituals}.


\subsection{Selecting Rituals}
\label{subsec:selecting_rituals}

Given the vast numbers of rituals that can fall into the six categories previously outlined, and the variance in the extents of their observance across cultures, another crucial decision is in selecting specific rituals as markers of cultural knowledge. There were several factors that we had to consider while making this decision. First and foremost, we needed activities whose identifying names are used in multiple cultures. For example, a ritual like {\em Passover} cannot be used since it is highly specific to Judaism and Jewish cultures. It would make no sense to ask {\em ``What are your cultural practices with regards to Passover?''} of a Hindu or a Muslim. Furthermore, the more specific the names, the easier it is for a QA system to associate its knowledge to that activity. As an example, {\em ``Catholic Christmas Mass''} is a highly denomination- and group-specific ritual, and will exhibit very little variation across cultures.
Second, in order to ease the data collection process, the selected rituals needed to have fairly concise and telegraphic names. For instance, it is confusing to probe a participant in a study about {\em ``the kinds of things you do before a sports game popular in your culture}; even though this is an activity that is fairly widespread and at the same time variable across cultures. Instead, we are seeking rituals that can be described in just a few words and bring a very specific activity or event to mind.
Our choice of rituals is intended to primarily ease the collection of crowd-sourced data---we thus pick activities that may have different practices across different cultural groups but are likely to be found in all of them. We take guidance from the analysis of \citet{bell1997ritual} and use the following six rituals as our target rituals in this work:


\begin{enumerate}
    \item {\bf Wedding}: wed, marriage, marry, matrimony, nuptials, wedlock, union, hymeneals ({\it rite of passage})
    \item {\bf Funeral}: funerary, burial, cremation, interment, entombment, obsequy ({\it rite of passage})
    \item {\bf Coming of Age}: becoming a man, becoming a woman, manhood, womanhood, adulthood ({\it rite of passage})
    \item {\bf Birth}: childbirth, delivery, birthing, childbearing, parturition, nativity ({\it rite of passage})
    \item {\bf New Year} ({\it calendrical rite})
    \item {\bf Birthday}: name day, natal day ({\it rite of passage})
\end{enumerate}

\noindent These rituals all have the advantage that across cultural groups there are limited number of ways of naming or expressing them, and the meaning is evident to most subjects answering our survey.


\section{Methodology}
\label{sec:methodology}

In this section, we outline the process of gathering the ritual-based cultural knowledge. We first describe the target cultures, followed by the details of our pilot experiment; we explain our data collection method on Amazon Mechanical Turk (MTurk); finally we describe the survey questionnaire that we used. We share these details both to explain our method as well as to enable reproducibility of the work.

\subsection{Selecting Target Cultures}
\label{subsec:selecting_target_cultures}

For this study, we focused on two specific target groups: Americans (people from the United States of America) and Indians (people from India). We chose these two groups for a variety of reasons. First, both the countries use English as one of their major languages, either officially or unofficially, and since our study was to be conducted in English, this was a key requirement. Second, Amazon Mechanical Turk has a high presence of workers from both these countries~\cite{difallah2018demographics}. Since the bulk of the data collection was to be done via Amazon's Mechanical Turk platform, it was essential that we considered the demographics of the crowd-workers. Moreover, these groups allowed us to set up a unique contrast and high degree of cultural variation between the two groups. Apart from providing contrasts against each other, the fact that the United States and India are large and diverse countries consisting of various cultures allows us to capture a varied amount of data within the groups themselves.

\subsection{Data Collection: Pilot Experiment}
\label{subsec:pilot_experiment}

For the pilot experiment, we collected data from a small number of participants for three rituals: \texttt{coming of age}, \texttt{wedding}, and \texttt{death rites/funeral}. We collected two unique sets of responses per ritual per culture, for a total of 12 unique responses. The identification of cultural group membership was done via self-identification by the participants, based on a demographic questionnaire that preceded the main survey. The participants took the survey using an online form. The actual survey consisted of a series of questions that were modifications of the ATOMIC~\cite{sap2019atomic} question set (see Section~\ref{subsec:survey_questionnaire} for details). The questions remained constant across rituals and cultures with only the initial prompt changing to keep the method as consistent as possible. The survey was conducted asynchronously and participants were compensated for their time.

\begin{figure}[h]
    \centering
    \includegraphics[width=0.4\textwidth]{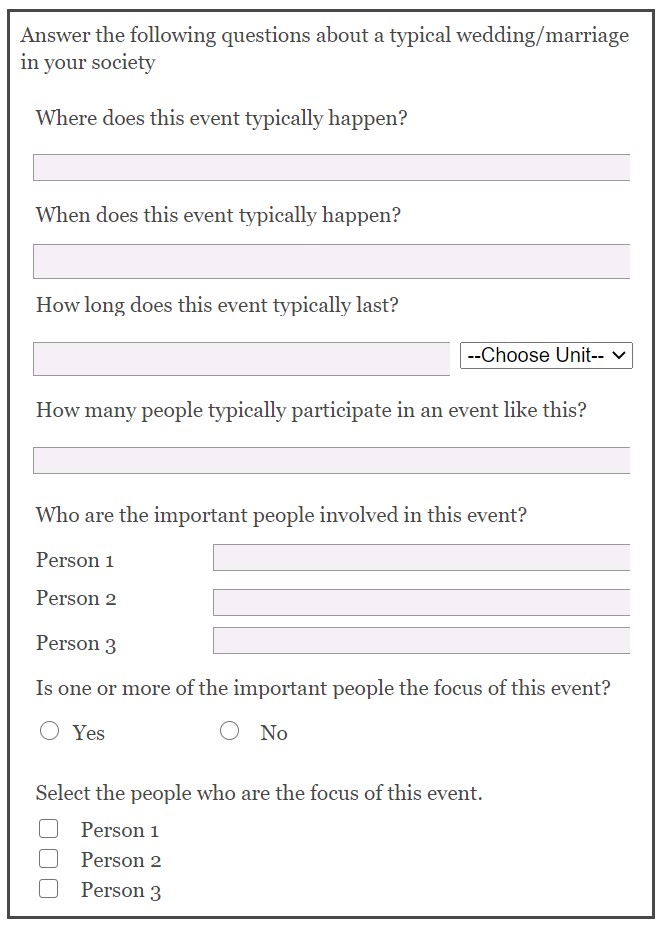}
    \caption{Layout of the main survey form as seen by the survey participants.}
    \label{fig:form_screenshot}
    \vspace{-5mm}
\end{figure}

\subsection{Data Collection: Amazon MTurk}
\label{subsec:mturk}

Once our small pilot study affirmed our idea, we moved ahead with collecting data on a slightly larger scale. We used Amazon Mechanical Turk (MTurk) to collect our second round of data. We set up separate tasks on MTurk for United States (US) participants and Indian (IN) participants, and geo-restricted the tasks to workers from the respective countries. We also restricted the tasks to \textit{Master} workers---workers with a work approval rate of $90\%$ or more. We set up a system of auto-generated survey codes that linked responses to MTurk workers without having to collect any Personally Identifiable Information (PII); these codes were used to filter out spam entries. The workers that provided spam responses were banned from the tasks. Overall, we collected a total of 32 useful data points for Indian participants, and 33 for US participants; these were divided roughly equally across rituals, with at least 5 responses per ritual.

\subsection{Survey Questionnaire}
\label{subsec:survey_questionnaire}

The survey questionnaire was divided into two sections: the self-identification questionnaire, and the survey proper. 

\subsubsection{Self-identification Questions}
\label{subsubsec:self_identification_questions}

We based the self-identification questionnaire primarily on the {\em College Students Knowledge and Belief} \cite{clark1981college} questionnaire. Portions of the questionnaire relevant to our current study were modified to suit our purposes, and the wording changed to be consistent with contemporary terminology. This self-identification questionnaire was further modified to prioritize the targeted countries with regard to language and religion. The options for these two fields were based on the most likely answers given the demographics of those populations. The updated questionnaire asked participants about their place of birth, languages used, and religion; it also asked these questions about their primary caregivers in order to ascertain the degree to which the participants were immersed into their identified culture. While for this work we have considered only nationality as the marker for culture, moving forward we would want to use a broader set of factors such as these to define the group.

\subsubsection{Event-specific Questions}

For the main survey, a prompt specifying the ritual event under consideration was first shown, and the participants were asked a set of questions pertaining to the specific event; see the questions asked and the layout of the form in Figure~\ref{fig:form_screenshot} for details. 





\subsubsection{Person-specific Questions}

After the event-specific questions, the participants were asked to answer questions that involved the specific persons mentioned in the events (as part of their responses). This part of the survey was adapted from ATOMIC~\cite{sap2019atomic}. Questions were divided into three temporal categories: before the event, during the event, and after the event. We considered four types of questions: (1) intent \& reaction; (2) need \& want; (3) effects; and (4) attributes. These four types evolved into 11 questions on the survey form, as shown in Figure~\ref{table:form_personx}. In terms of the presentation of the questions, the terms \texttt{PersonX} or \texttt{PersonY} were replaced by the actual names that participants provided (in the Person fields) in order to make the questions feel more natural to the survey participants. These person-specific questions were repeated for each person that the survey participant deemed ``important'' to a given event. 

\begin{table}
    \centering
    \framebox[3.5in][c]{\rule{0pt}{1.5in}
    \noindent\begin{tabular}{p{1\linewidth}}
	\textbf{\textsc{Before the event}} \\
	\midrule
	Does this person typically have an intent in causing the event?\\
	What is this person's typical intent in causing the event? \\
    Does this person typically need to do anything before this event? \\
    What does this person typically need to do before this event? \\
	\midrule
    \textbf{\textsc{During the event}} \\
	\midrule
    Does this person typically use something during the event? \\
    What things does this person typically use during this event? \\
	\midrule
    \textbf{\textsc{{After the event}}} \\
	\midrule
    How would this person be described as a consequence of the event? \\
    Does this person typically want to do something after this event? \\
    What does this person typically do after this event? \\
    What is the typical effect of the event on this person? \\
    What does this person typically feel after this event? \\
    \end{tabular}
    }
    \caption{Questions asked of the survey participants for each \textit{important} person associated to an event in their responses.}
    \label{table:form_personx}
    \vspace{-5mm}
\end{table}





\section{Results}
\label{sec:results}

In this section, we detail the findings of our study. We first report on the demographics of the survey participants, particularly with an eye towards cultural background, and then recount and discuss the responses to the survey.

\subsection{Demographics}
\label{subsec:results_demographics}

We obtained a total of 77 unique survey responses with each participant limited to one response per ritual per culture. Only information that was deemed necessary for the purposes of the experiment was collected from participants, with an aim to avoid collecting PII as far as possible.

\subsubsection{Geography}
\label{subsubsec:geography}

Out of the 77 responses, 38 were from India and 39 from the US. While all the participants that identified as American (from the US) resided in the US; in the case of Indian participants, three people said they currently resided in the US. 


\begin{figure}[h]
\centering
\subfloat[Subfigure 1 list of figures text][Religion Data for the participants from India.]{
\includegraphics[scale=0.64]{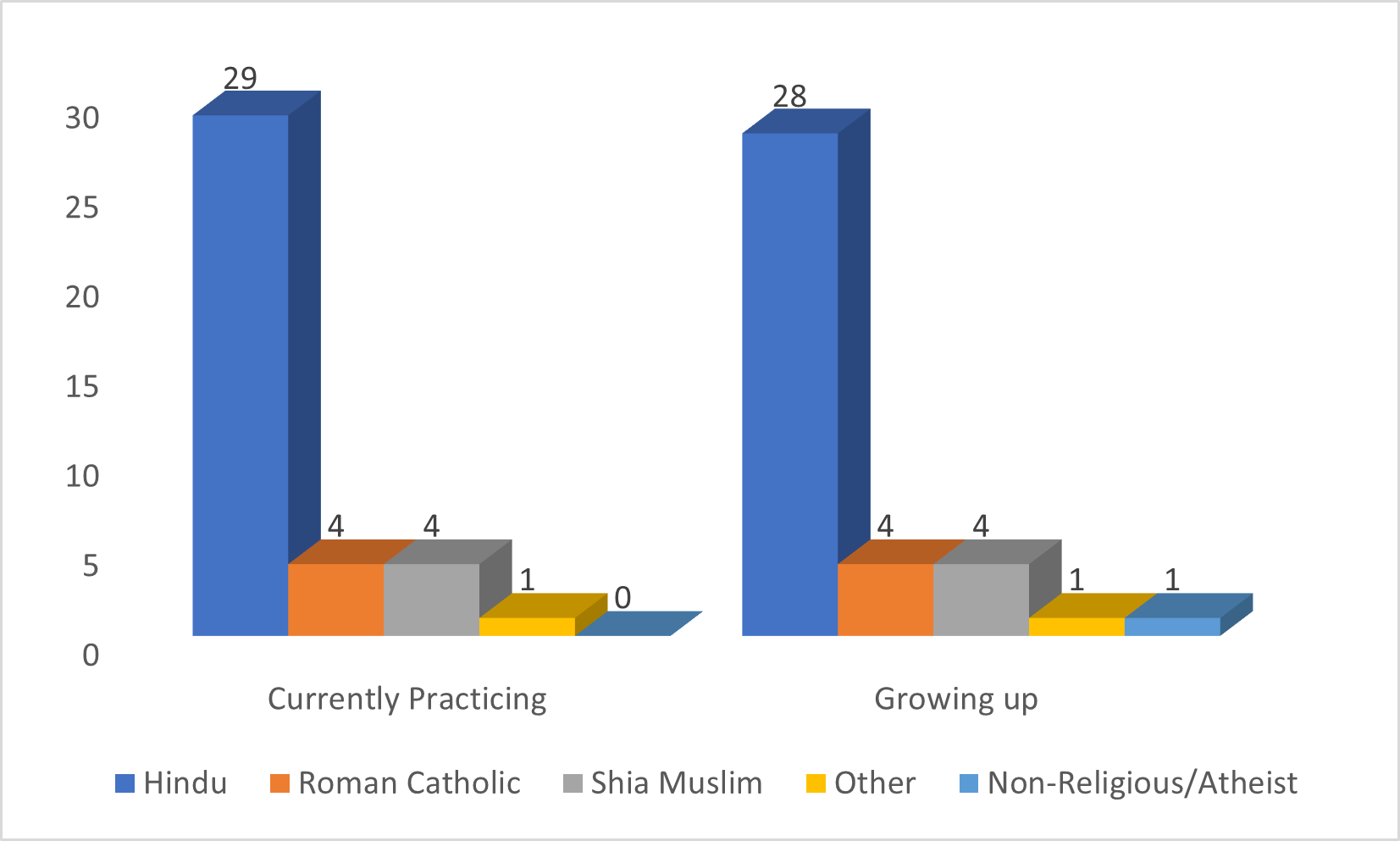}
\label{fig:religion_demographics_in}}
\qquad
\\
\subfloat[Subfigure 2 list of figures text][Religion Data for the participants from the US.]{
\includegraphics[scale=0.64]{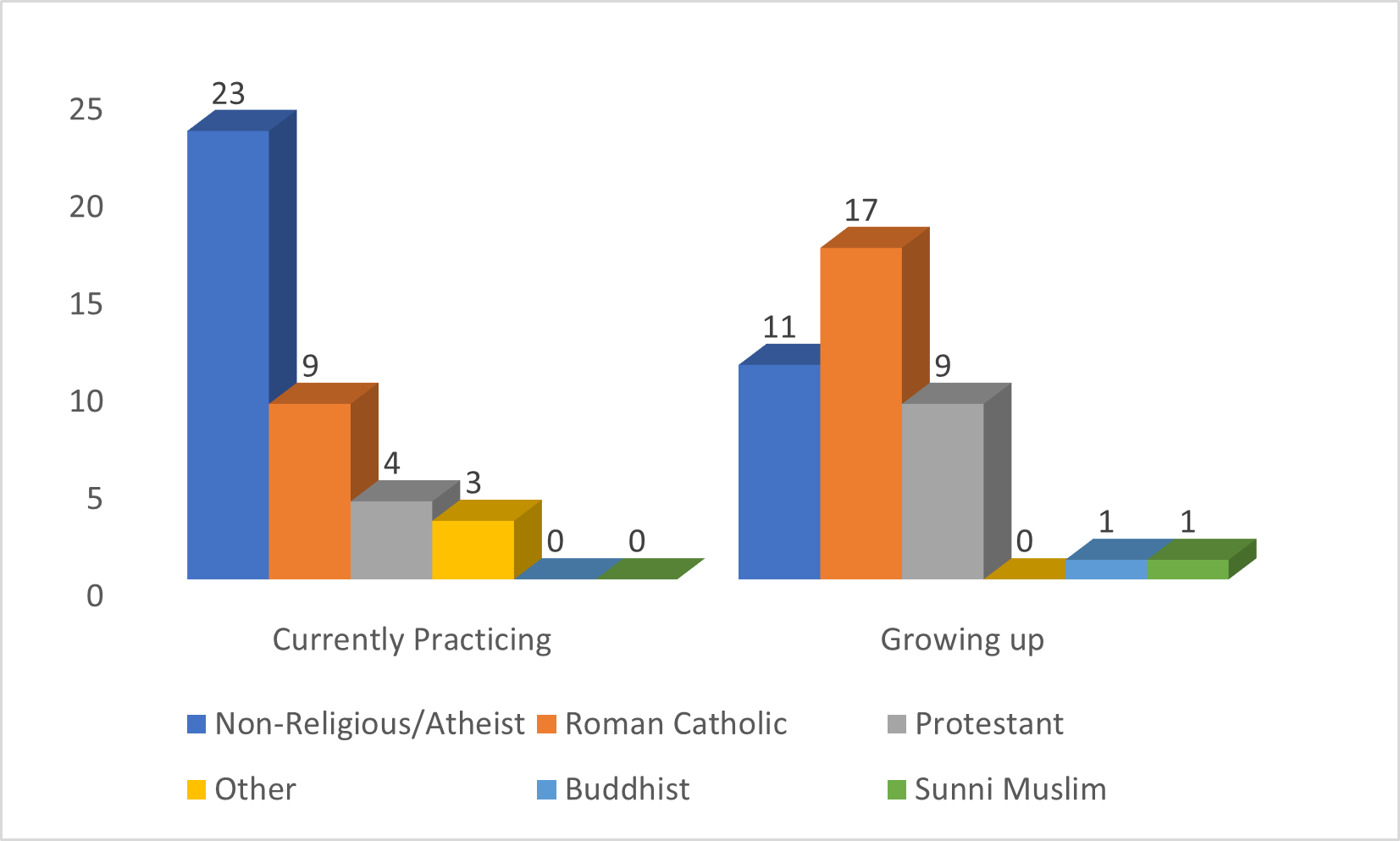}
\label{fig:religion_demographics_us}}
\caption{Variation in the religion practiced by participants growing up versus currently for both cultures.}
\label{fig:demographics}
\vspace{-2mm}
\end{figure}

\subsubsection{Religion}
\label{subsubsec:religion}

Since many rituals have a basis in religion---to the extent that they are often intertwined~\cite{goody1961religion, geertz1966religion, bell1992ritual}---it is important to ensure a diversity of religious practice among the respondents. Of the participants in the study, 29 said they practiced Hinduism; 17 said they practiced some form of Christianity; four Islam; four other religions; and 23 said they did not practice any religion and/or were atheist ($N=77$). Among participants who identified as Indian, there was significantly less variation in religion compared to the participants who identified as Americans. There was also significantly more change in religion practiced over a lifetime for US participants as compared to IN. This is illustrated in Figure~\ref{fig:demographics}.

\subsubsection{Language}
Among the $N=77$ participants, 39 identified their native language as English; 17 Tamil; seven Hindi; five Telugu; four Urdu; and one each as five other languages.


\subsection{Qualitative Analysis of Survey Responses}

Over both the pilot and MTurk studies, we collected a total of 77 responses, with a minimum of 5 responses per ritual (for six rituals, see Section~\ref{subsec:selecting_rituals} for a list) per cultural group. Since there are not enough data points yet in our pilot experiment to conduct a quantitatively significant study, we instead present a qualitative analysis of the data collected thus far.

\begin{figure}[h]
\centering
\includegraphics[width=0.4\textwidth]{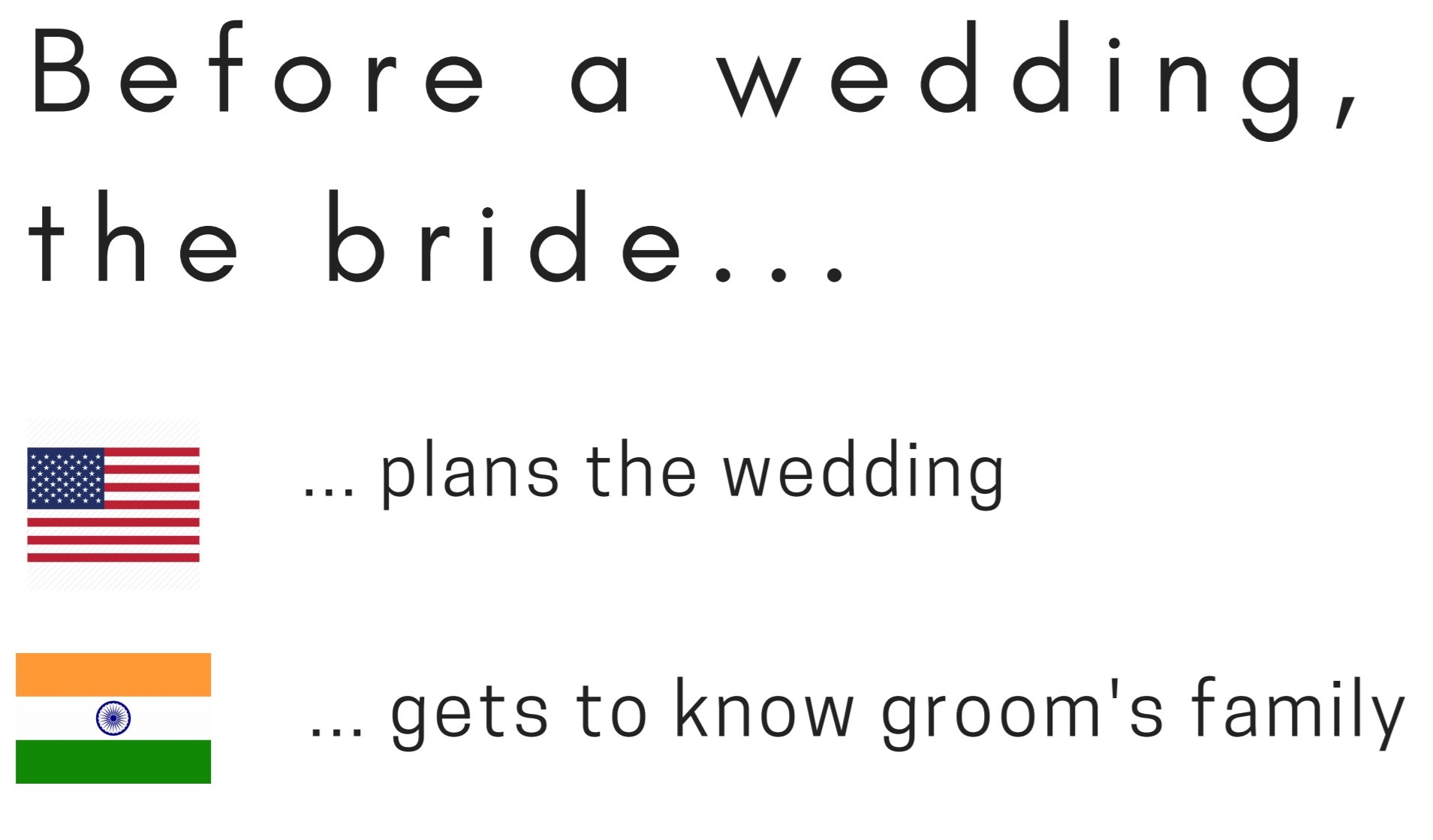}
\caption{The differences in expectations of a bride in the US versus in India.}
\label{fig:bride}
\end{figure}

The results show us that while rituals have some common features across cultures, they can also have significant variations to the point where the {\it common knowledge} would be noticeably different. Let us look at one significant difference seen in the responses: for the {\it wedding} ritual, participants from the US said the bride would focus on the wedding planning part of the event, like dresses and so forth; while the Indian participants focused on the cultural aspects of the wedding, as well as the fact that the bride might have to get to know the groom's family, and possibly the groom himself, as illustrated in Figure~\ref{fig:bride}. It would be extremely unlikely for a bride to not know the groom's family, let alone the groom himself, in a typical US wedding; while this is still reasonably prevalent in Indian society. This is an excellent example of the type of knowledge that is collected by our work, where a machine can now leverage this information as commonsense knowledge that is culturally sensitive and correct. This also suggests that with more data, more such variations in the way rituals are conducted across cultures can be documented and understood by NLP systems.

While the data showed that the person-roles involved in rituals were more or less similar across cultures, the order of importance indicated was noteworthy, and shown in Table \ref{tab:ritual_key_people}. For birth rituals, participants in the US considered the birth itself the main event, and said the Doctor (physician) was an important person (after the parents); while Indian participants focused more on the family. It is also worth noting that US participants put Mother before Father, while Indian participants reversed the priority order. All participants gave similar responses in terms of the important person-roles for a wedding: the bride, the groom, and their parents. An important difference was that Americans identified the Bride's mother as the more important person, while Indians picked the Groom's mother for the same. Likewise, for a funeral, more priority was given to the pastor or priest by US participants; unlike Indian participants, who considered the family to be the more important participants.

\begin{figure}[h!]
\centering
\subfloat[Subfigure 1 list of figures text][Duration of each ritual in hours.]{
\includegraphics[scale=0.65]{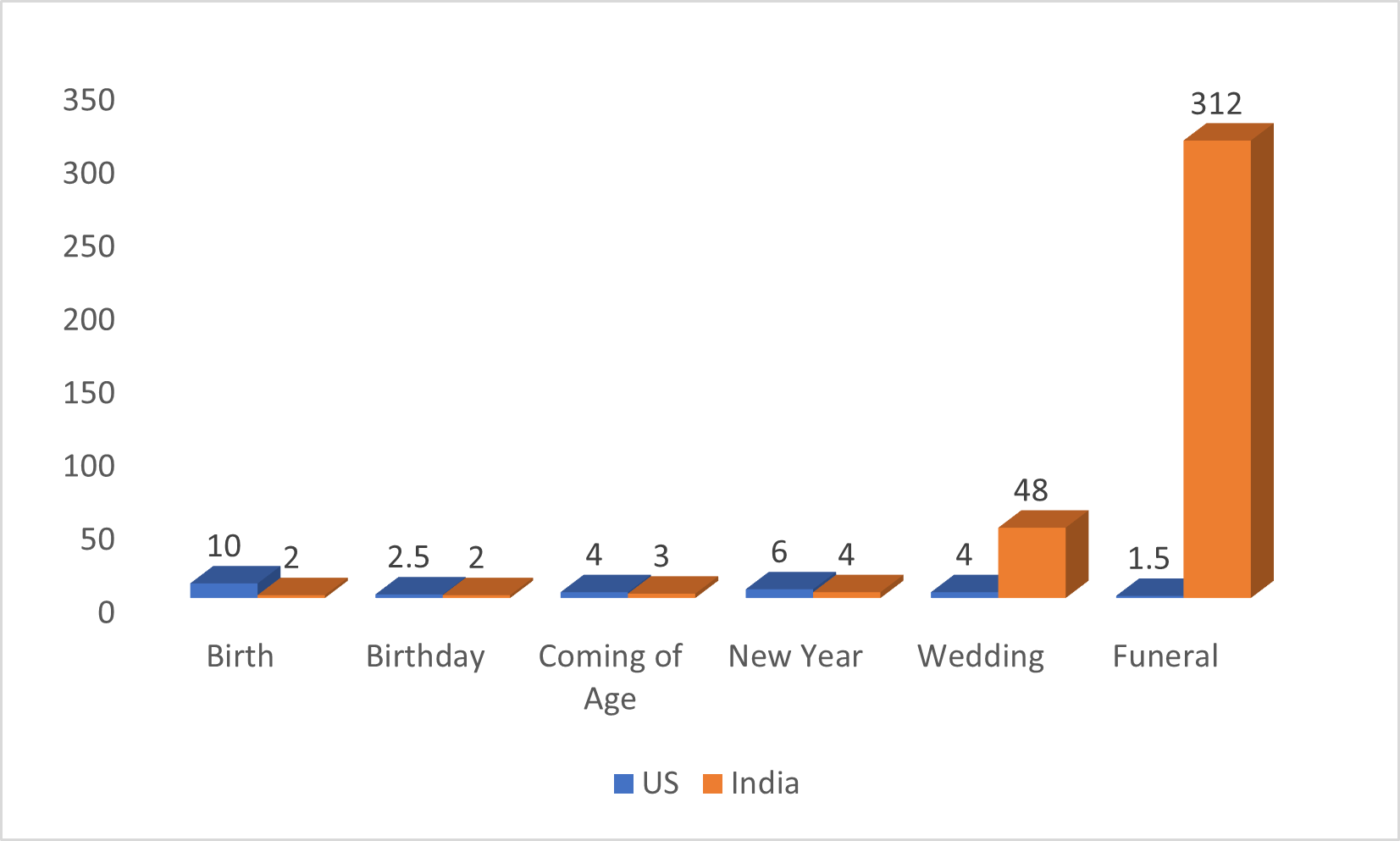}
\label{fig:ritual_duration}}
\qquad
\\
\subfloat[Subfigure 2 list of figures text][Number of people in each ritual.]{
\includegraphics[scale=0.65]{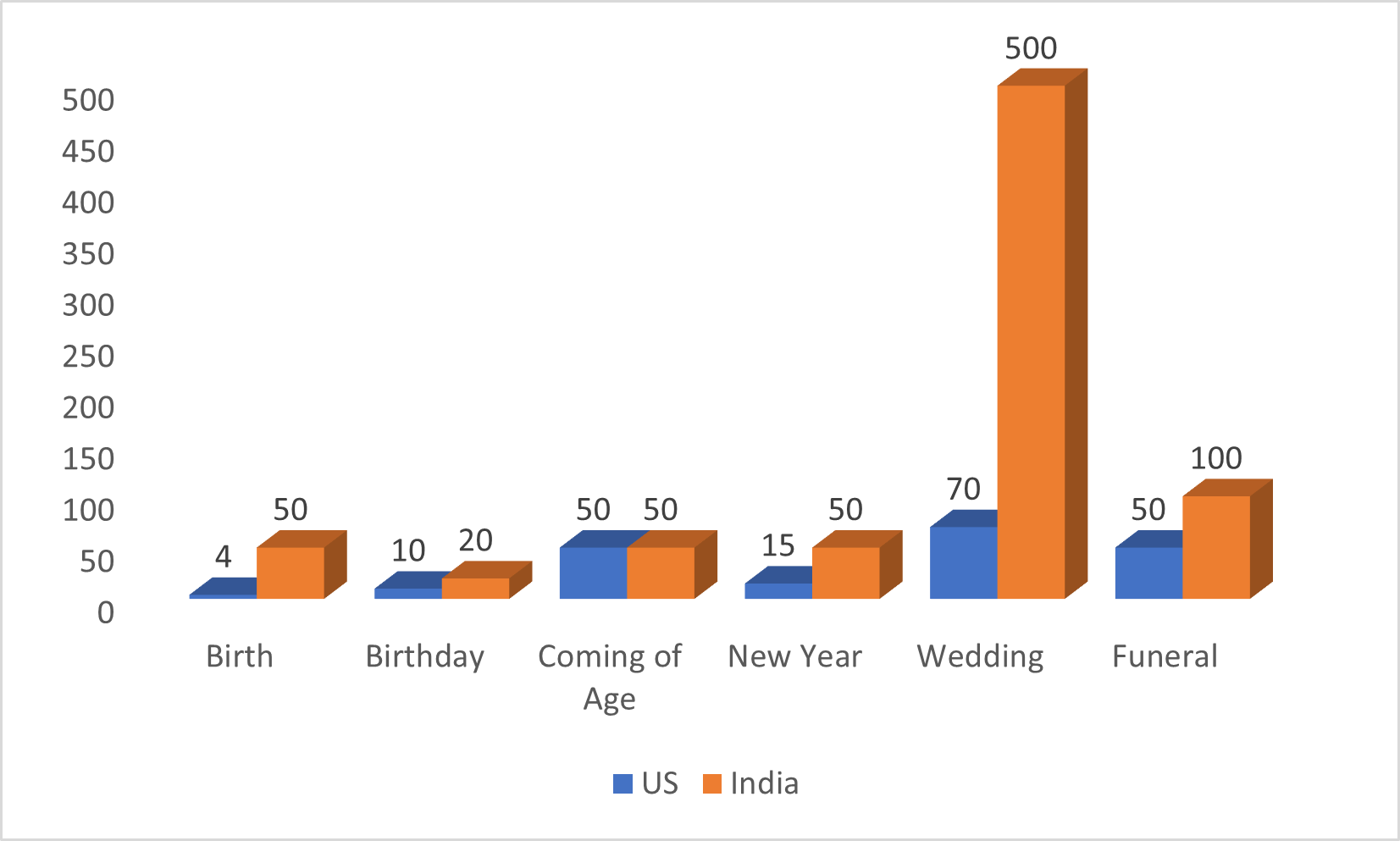}
\label{fig:ritual_participation}}
\caption{The variations in the duration and number of participants by culture (\textit{The median is taken for each case}).}
\label{fig:ritual_response}
\vspace{-2mm}
\end{figure}

Another set of major differences were seen in terms of the durations of the rituals. While the duration of the other four rituals were more or less comparable, weddings and funerals had significant differences across cultures. For the {\it wedding} ritual, participants from the US said that the ritual typically lasted a few hours, while Indian participants responded by saying that weddings lasted multiple days. The difference was even more striking for funerals, which US participants said lasted only a few hours; while Indian participants reported funeral rites lasted 13 days, as shown in Figure~\ref{fig:ritual_response}(a).

One other key difference observed was the number of people that participated in each ritual. With the exception of {\it coming-of-age} rituals, for which the number of participants were reported to be similar, the rituals in India seemed to involve a lot more people than in the US. While the difference in numbers was already pronounced for other rituals (Indian funerals seem to have {\it twice} as many people as US ones), the striking difference is weddings, where Indian ones had several hundred guests, while US ones remained below 100, as shown in Figure \ref{fig:ritual_response}(b).

These findings validate our expectation that rituals can give us a peek into cultures and how they vary, and that commonsense knowledge cannot truly be complete without including cultural nuances.

\section{A Cultural Commonsense Graph}
\label{sec:construct_graph}



\begin{figure}[h]
\centering
\subfloat[Subfigure 1 list of figures text][India]{
\includegraphics[width=0.5\textwidth]{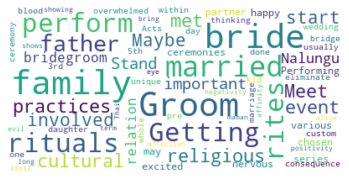}
\label{fig:wedding_response_in}}
\qquad
\\
\subfloat[Subfigure 2 list of figures text][USA]{
\includegraphics[width=0.5\textwidth]{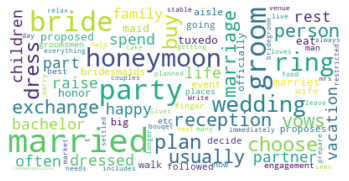}
\label{fig:wedding_response_us}}
\caption{Most common words in the responses for \textit{wedding} ritual for both cultures represented in word clouds.}
\label{fig:wordclouds}
\vspace{-2mm}
\end{figure}

The main motivation behind this study is to eventually distill the knowledge thus gathered into a form useful to current NLP systems. Specifically, our hope is to transform the cultural knowledge thus collected into a resource that can be used to produce more human-like performance in NLP tasks. Past efforts at systematizing commonsense knowledge~\cite{sap2019atomic,bosselut2019comet,speer2016conceptnet,liu2004conceptnet} for the NLP community's use have taken a similar path, starting from data collection through analysis and summarization of the data, then eventually the construction of knowledge graphs from those summaries. For our work, these summaries look like the examples shown in Figure~\ref{fig:wordclouds}. Specifically, we consider the {\it wedding} ritual for both US and IN cultures. A quick examination of Figure~\ref{fig:wordclouds}(a) shows that while for IN, there seems to be a significant emphasis on the religious \& ritual aspects as well as the family of the participants; whereas for US, shown in Figure~\ref{fig:wordclouds}(b), the emphasis is much more on the celebration as a {\em party}, and the social aspect of the event.

\begin{figure*}[h]
    \centering
    \includegraphics[width=1.0\textwidth]{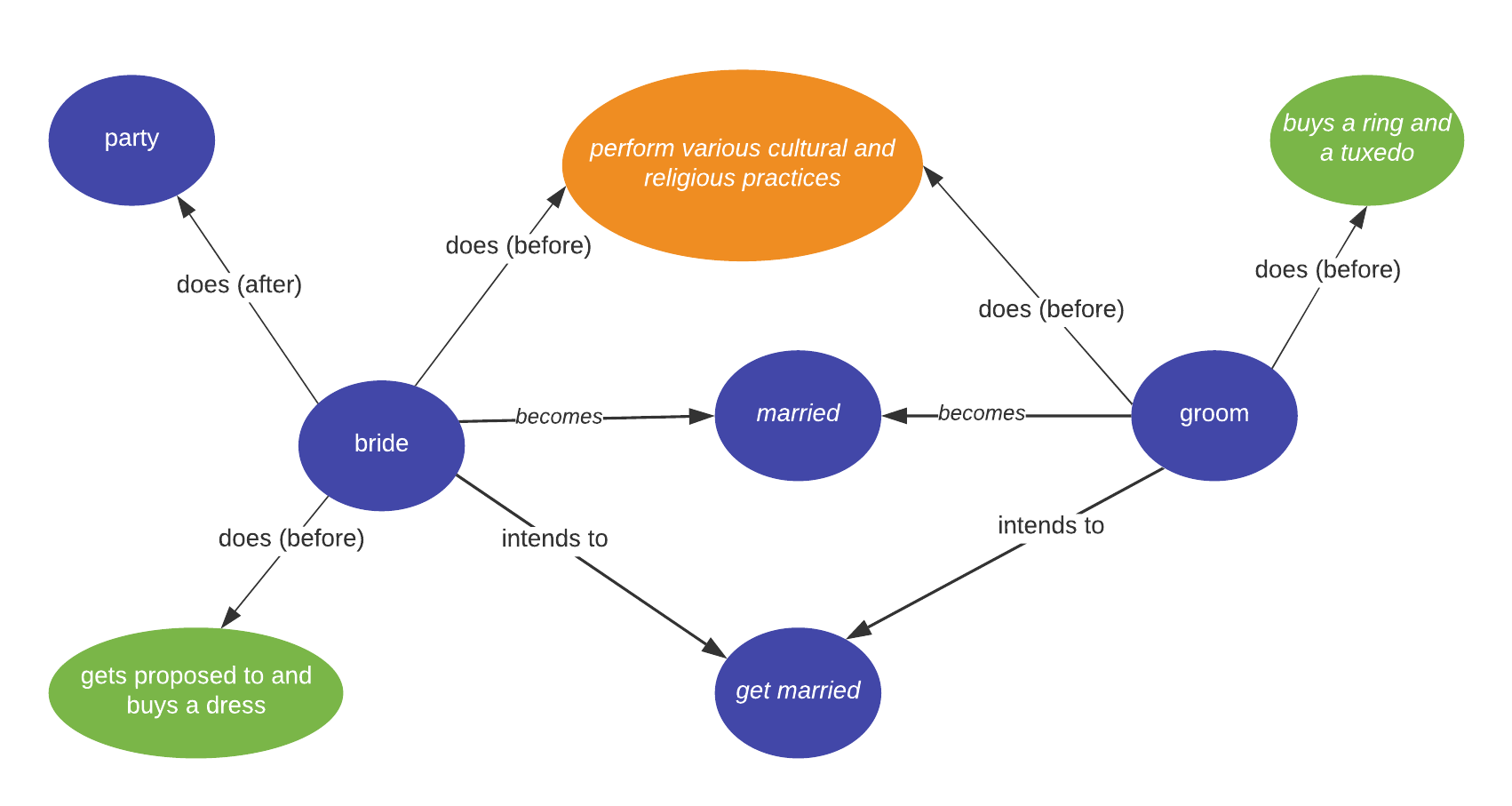}
    \caption{A graph for the subset of the data collected for the {\it wedding} ritual. The nodes in orange show responses of Indian participants; those in green show responses of US participants; nodes in purple are common to both. The thickness of the edge indicates the frequency of the relationship. The figure has been manually edited for grammar, and is intended for illustration.}
    \label{fig:cskg_example}
    \vspace{-5mm}
\end{figure*}

The next step in our process is to represent this abstract space of words in the form of a directed graph composed of entities and relations, akin to most existing knowledge graphs. In order to do this, we performed several stages of cleaning and data filtering of the data to extract the relevant sections, followed by aggregating the details present in the data for each {\it event} and {\it person} together. Case-folding was done on the nodes and edges to avoid duplicates. After this, we extracted all potential nodes from each instance of data, with the relevant {\it event} or {\it person} being the origin node; the prompt text as seen in Table~\ref{table:form_personx} and Figure~\ref{fig:form_screenshot} as the (relation) edge, while the answer for each question was the corresponding destination node. This data was then visualized as a network using Python's GraphViz package \cite{ellson2004graphviz}, merging nodes with the exact same labels into a single node, with nodes corresponding to the two target countries being assigned different colors. An illustrative example of the end-result of this process is shown in Figure~\ref{fig:cskg_example}. This sub-graph is a glimpse of what our entire knowledge base will look like after we run entire dataset through a rigorous NLP pipeline containing semantic role labeling, word sense disambiguation, syntactic and semantic parsing, among others. This sub-graph and the larger graph that it is a part of---which can also be represented in the traditional adjacency graph format for consumption by NLP systems---are the ultimate goal of our work on creating a cultural commonsense knowledge graph. We are unable to present more examples in this manuscript due to space limitations, however, we have collated a representative sample of our efforts thus far, including the current set of networks from the process above, at the following URL: \textcolor{blue}{\url{rebrand.ly/atlas-graph}}.

\section{Conclusion \& Future Work}
\label{sec:future_work}


We have presented the setup and results of a pilot experiment collecting cultural information from diverse groups about different life rituals. We seek to create a repository of cultural commonsense knowledge. We envision that such knowledge could greatly improve the ability of AI systems to exhibit human-like performance by addressing gaps in their current knowledge. We showcased some interesting qualitative results. The task of injecting cultural sensitivity into commonsense reasoning, while being crucial to developing a true human-like AI, has not been previously explored in the field. We exposed this gap and in order to bridge it performed the difficult task of choosing suitable cultural markers that would work within existing frameworks of commonsense knowledge. Moving forward, it is important that we carefully assess how culturally variable different events are, as well as measure the significance of any differences for these events across culture as we look to scale up this work. Another direction for future work is to explore the various knowledge representation techniques to find the best way to represent what will likely be multi-dimensional data. While the data we have collected thus far is too small to be used to directly improve the performance of QA or other NLP tasks, the approach here allows for scale-up into a full dataset. We are currently in that process, and we are also evaluating existing datasets to design benchmarks that require the use of  cultural commonsense.
\section{Acknowledgements}
This project was supported in part by an 2019 IBM Career Award to Prof. Finlayson, as well as in part by DARPA contract FA8650-19-C-6017. 

\small
\bibliography{paper}

\newpage
\section*{Appendix A: Ethical Considerations}
~\\
We were extremely cognizant of the ethical aspects of our study. The details of the study were reviewed by the Institutional Review Board (IRB) of Florida International University. The proposed study was deemed \textit{Exempt Research}, and given approval to proceed. No step of the study---excepting internal preparations---were carried out prior to obtaining this approval. No personally identifiable information was collected from the participants, and all IRB requirements were met and followed during the course of the study.

Another issue is whether a study such as this might exacerbate existing stereotypes. We agree that this is a concern, and requires careful consideration in future work.


\section*{Appendix B: Additional Figures and Tables}

\begin{table}[h]
\begin{tabular}{p{0.20\linewidth}p{0.20\linewidth}p{0.20\linewidth}p{0.20\linewidth}}
    \toprule
	\textbf{\textit{US}} & \textbf{\textit{India}} & \textbf{\textit{US}} & \textbf{\textit{India}}\\
	\midrule
	\multicolumn{2}{c}{\textbf{Birth}} & \multicolumn{2}{c}{\textbf{Birthday}}\\
	\midrule
	Mother & Father & Self & Parents\\
	Father & Mother & Partner & Family\\
	\textbf{\textit{Doctor}} & \textbf{\textit{Family}} & Family & Relatives\\
	\midrule
	\multicolumn{2}{c}{\textbf{New Year}} & \multicolumn{2}{c}{\textbf{Coming-of-age}}\\
	\midrule
	Spouse & Friends & Self & Parents\\
	Parents & Spouse & Parents & Family\\
	Friends & Family & Siblings & Relatives\\
	\midrule
	\multicolumn{2}{c}{\textbf{Wedding}} & \multicolumn{2}{c}{\textbf{Funeral}}\\
	\midrule
	Bride & Bride & Spouse & Father\\
	Groom & Groom & \textbf{\textit{Pastor/Priest}} & Son\\
	\textbf{\textit{Bride's Mother}} & \textbf{\textit{Groom's Mother}} & Parents & Family\\
	\bottomrule
\end{tabular}
\caption{Important people for each ritual by culture.} 
\label{tab:ritual_key_people}
\end{table}

\label{subsec:results_survey_responses}

\end{document}